\documentclass[letterpaper, 10 pt, journal, twoside]{IEEEtran}
\ifCLASSINFOpdf
\else
\fi
\usepackage{times}
\usepackage{epsfig}
\usepackage{graphicx}
\usepackage{caption}
\usepackage{subcaption}
\usepackage{amsmath, mathtools}
\usepackage{amssymb}
\usepackage{authblk}
\usepackage{cancel}
\usepackage{mdframed}
\usepackage{booktabs}
\usepackage{comment}
\usepackage{xcolor}
\usepackage{pifont}
\newcommand{\cmark}{\ding{51}}%
\newcommand{\xmark}{\ding{55}}%

\newcommand{\ignore}[1]{}

\newcommand{\ba}{\begin{array}}
\newcommand{\ea}{\end{array}}
\newcommand{\bc}{\begin{center}}
\newcommand{\ec}{\end{center}}
\newcommand{\be}{\begin{enumerate}}
\newcommand{\ee}{\end{enumerate}}
\newcommand{\bea}{\begin{eqnarray}}
\newcommand{\eea}{\end{eqnarray}}
\newcommand{\beas}{\begin{eqnarray*}}
\newcommand{\eeas}{\end{eqnarray*}}
\newcommand{\beq}{\begin{equation}}
\newcommand{\eeq}{\end{equation}}
\newcommand{\bfig}{\begin{figure}}
\newcommand{\efig}{\end{figure}}
\newcommand{\bi}{\begin{itemize}}
\newcommand{\ei}{\end{itemize}}
\newcommand{\bpic}{\begin{picture}}
\newcommand{\epic}{\end{picture}}
\newcommand{\btabular}{\begin{tabular}}
\newcommand{\etabular}{\end{tabular}}
\newcommand{\btable}{\begin{table}}
\newcommand{\etable}{\end{table}}

\newcommand{\es}{\vfill
                 \rule[-6mm]{170mm}{0.7mm} \\
                 \redw{{\tiny
		  \hfill S-\theslide}}
                 \end{slide}}

\newcommand{\matxx}[1]{{\mathbf #1}}
\newcommand{\vecXX}[1]{{\mathbf {#1}}}
\newcommand{\vecYY}[1]{{\boldsymbol {#1}}}

\newcommand{\argmax}{\operatornamewithlimits{arg\ max}}

\def \hbar {{\bar{h}}}

\def \xbar {{\bar{x}}}

\def \vect {{\vecXX{t}}}

\def \vecx {{\vecXX{x}}}

\def \vecz {{\vecXX{z}}}

\def \veceta   {{\vecYY{\eta}}}
\def \vecmu    {{\vecYY{\mu}}}

\def \matB {{\matxx{B}}}

\def \matI {{\matxx{I}}}
\def \matJ {{\matxx{J}}}

\def \matT {{\matxx{T}}}

\def \matW {{\matxx{W}}}

\def \matSigma  {{\matxx{\Sigma}}}
\def \matLambda {{\matxx{\Lambda}}}

\def \deg {^{\circ}}

\makeatletter
\renewcommand*\env@matrix[1][*\c@MaxMatrixCols c]{%
  \hskip -\arraycolsep
  \let\@ifnextchar\new@ifnextchar
  \array{#1}}
\makeatother

\newcommand{\RR}{\mathbb{R}}

\newcommand{\SE}[1]{\ensuremath{\mathbf{SE}(#1)}}

\newcommand{\SO}[1]{\ensuremath{\mathbf{SO}(#1)}}

\font\Bigmath=cmsy10 scaled \magstep2
\def\dplus{\mathrel{%
  \ooalign{$+$\cr\hss\lower.255ex\hbox{\Bigmath\char5}\hss}}}  
\def\dminus{\mathrel{%
  \ooalign{$-$\cr\hss\lower.255ex\hbox{\Bigmath\char5}\hss}}}

\newcommand{\cM}{\mathcal{M}}
\newcommand{\cN}{\mathcal{N}}

\newcommand{\cU}{\mathcal{U}}

\newcommand{\cX}{\mathcal{X}}
\newcommand{\cY}{\mathcal{Y}}

\DeclareMathOperator*{\Exp}{Exp}
\DeclareMathOperator*{\Log}{Log}

\def\bea#1\eea{\begin{align}#1\end{align}}
\def\beas#1\eeas{\begin{align*}#1\end{align*}}
   
\usepackage[pagebackref=true,breaklinks=true,colorlinks,bookmarks=false]{hyperref}

\begin{document}
%
\title{Distributed Simultaneous Localisation and Auto-Calibration using Gaussian Belief Propagation
}
%
%
%

\author{Riku Murai, Ignacio Alzugaray, Paul H.J. Kelly and Andrew J. Davison%
\thanks{Manuscript received: July, 14, 2023; Revised November, 20, 2023; Accepted December, 18, 2023.}
\thanks{This paper was recommended for publication by Editor Sven Behnke upon evaluation of the Associate Editor and Reviewers' comments.
This work was supported by EPSRC (EP/K008730/1 and EP/P010040/1).} 
\thanks{Authors are with Department of Computing, Imperial College London, UK 
        {\tt\footnotesize \{riku.murai15, i.alzugaray, p.kelly, a.davison\}@imperial.ac.uk}}%
\thanks{Digital Object Identifier (DOI): see top of this page.}
}
%
%

\markboth{IEEE Robotics and Automation Letters. Preprint Version. Accepted January, 2024}
{Murai \MakeLowercase{\textit{et al.}}: Distributed Simultaneous Localisation and Auto-Calibration using GBP} 

%



\maketitle

\begin{abstract}
We present a novel scalable, fully distributed, and online method for simultaneous localisation and extrinsic calibration for multi-robot setups.
Individual a priori unknown robot poses are probabilistically inferred as robots sense each other while simultaneously calibrating their sensors and markers extrinsic using Gaussian Belief Propagation.
In the presented experiments, we show how our method not only yields accurate robot localisation and auto-calibration but also is able to perform under challenging circumstances such as highly noisy measurements, significant communication failures or limited communication range.
\end{abstract}


%
\IEEEpeerreviewmaketitle

\section{Introduction}
%
%
%
%
\IEEEPARstart{A}{s} robots become increasingly ubiquitous, more of them are expected to jointly operate and coordinate in shared environments to perform complex and collaborative tasks.
For many of the tasks, it is critical to estimate the robot's pose accurately and precisely, also known as robot localisation.
While many localisation systems have been proposed in the literature, most of the works focus on single-robot setups.
However, relying solely on a single robot's proprioceptive and exteroceptive sensors for localisation is challenging and often limiting, especially in a multi-robot scenario, where relative localisation is crucial to ensure that robots interact safely and effectively.

This motivates {\em multi-robot collaborative localisation}, where robots utilise each other's observations to enhance their own localisation accuracy and therefore, the overall accuracy of the multi-robot system. 
The accuracy of co-localisation, however, is heavily reliant on the quality of extrinsic calibration of the sensors (e.g. visual camera rigs, rangefinders) and the markers they can detect on other robots (e.g. AprilTags, reflective markers).
While most works often take such extrinsic calibration for granted,
in practice, the default in-factory calibration can only be precise to a certain degree.
This is particularly important in multi-robot setups, where manual calibration becomes impractical and highly accurate in-factory, per-robot calibration incurs high operational costs.

\begin{figure}[!tbp]
  \center
  \includegraphics[width=\linewidth]{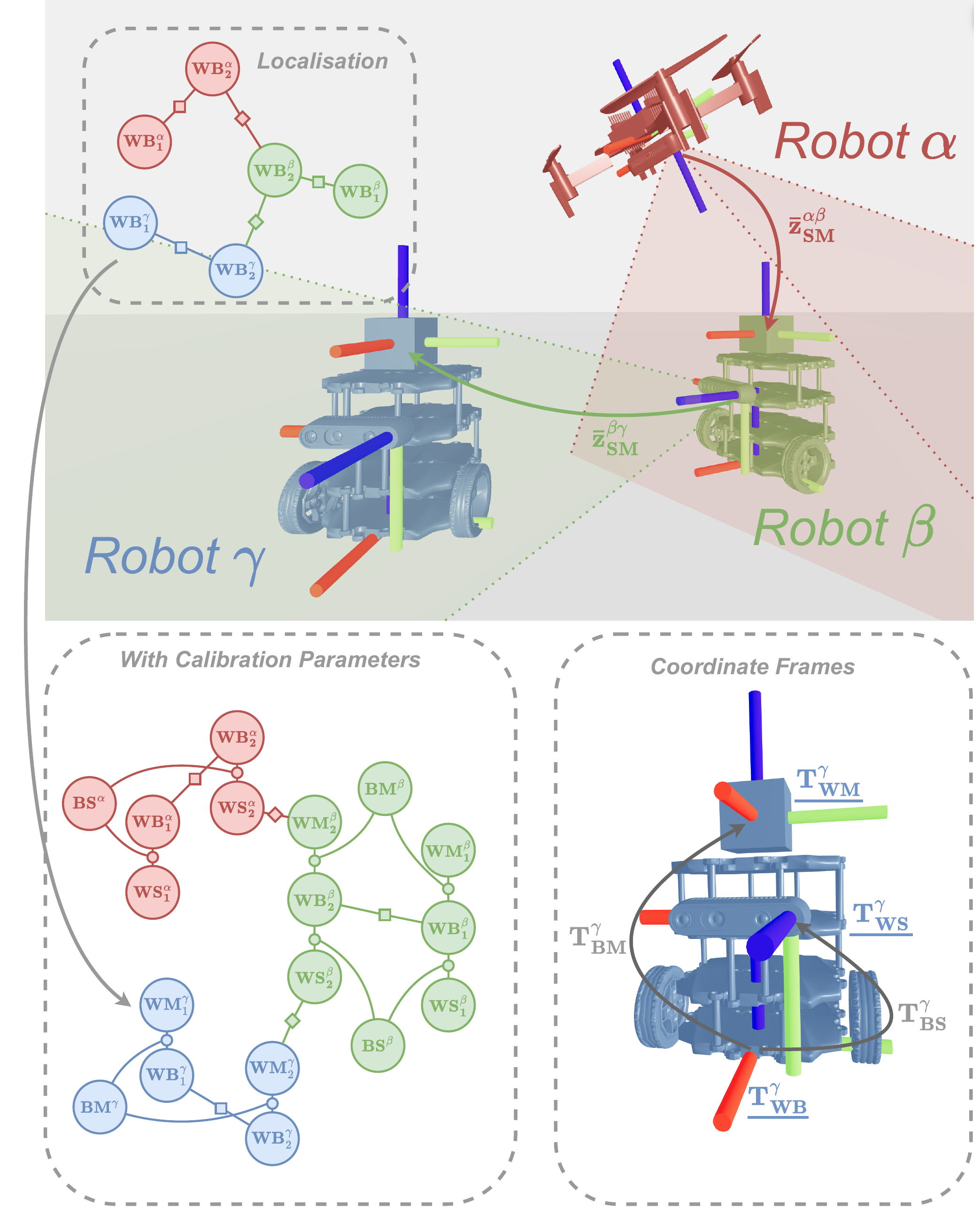}
  \caption{
Overview of the proposed auto-calibrating localisation system for three heterogeneous robots (top). 
Each robot observes the markers $M$ placed on its peers to establish measurement $\bar{\vecz}_{SM}$
using sensor $S$ mounted on top of a moving base $B$. 
Using the proposed methodology, the robots' relative positions and their calibration parameters can be retrieved in a distributed and asynchronous fashion performing probabilistic inference on a factor graph.
We refer to $\matT_{WB}$ as $\matW\matB$ for clarity.  }\label{fig:full-calibration-graph}
\vspace{2mm} \hrule
\vspace{-2em}
\vspace{2mm}
\end{figure}

In this paper, we envision a system (see Fig.~\ref{fig:full-calibration-graph}) in which multiple robots co-localise themselves as they move and sense each other while simultaneously estimating and refining the extrinsic calibration of their sensor and their onboard marker on-the-fly. 
In summary, the contributions of our work are:
\bi
\item A novel method for distributed multi-robot localisation and extrinsic calibration of both the sensor and the observed marker on the robots. Our approach builds on top of our previous work {\em Robot Web}~\cite{Murai:etal:TRO:2023}, originally limited to pose variables in \SE2 given range-bearing observations. We extend the framework by simultaneously estimating the \SE3 pose of the robots and their extrinsic calibrations using Gaussian Belief Propagation (GBP) to further improve the accuracy of multi-robot localisation.
\item We present a formulation of the inter-robot factor that avoids the sharing of the calibration variables amongst multiple robots, sparing communication effort between robots and thus enhancing the scalability of the system.
\item We provide an extensive evaluation of our approach in comparison with other state-of-the-art alternatives and measure the performance of the method under extreme conditions such as a large number of communication failures, a large proportion of outlying measurements, and a limited communication range.
\ei

%

\section{BACKGROUND}
\subsection{Multi-Device Calibration}
Calibration is vital for robotic operations and, as such, the body of literature on the subject is vast. 
Due to space limitations, our literature review focuses on calibration processes that involve multiple cameras or multiple robots.

Accurate extrinsic calibration is often critical in multi-device systems. Different robots have different base frames, and within a robot, the exact position of the sensor and observable onboard marker may not be available. 
A common instance of this is hand-eye calibration. 
From a set of known relative transformations, the calibration process seeks to establish the undetermined relationship, often the relative transformation between the robot base frame and sensor frame~\cite{Shiu:Ahmad:RAL1989}.
These methods can be extended to support multiple robots using iterative methods~\cite{Wang:etal:IROS2014} or probabilistic approaches~\cite{Ma:etal:AR2018}.
However, these methodologies primarily focus on offline settings where calibration precedes operational activities.
In multi-robot setups,~\cite{Gowal:etal:IROS2011} proposes a method to perform online calibration of infra-red sensors while estimating the parameters of the underlying physical sensor model.

Multi-camera rigs are becoming increasingly popular as they can significantly extend the surrounding perceptive field for any robot and even directly yield stereo-depth capabilities provided there is view overlap.
However, accurate calibration of such these rigs is often challenging, leading to several works on automatic offline calibration~\cite{Esquivel:etal:PR2007,Carrera:etal:ICRA2011,Yukai:etal:ECCV2020}.
The method in~\cite{Dang:etal:TIP2009} carries out continuous self-calibration using an extended Kalman filter in a stereo setup.
Beyond two cameras, self-calibration of multiple cameras extrinsic is achieved on an aerial vehicle in~\cite{heng:etal:AR2015}, whereas an information-theoretic approach described in~\cite{Dexheimer:etal:RAL2022} is able to operate on a rig of eight cameras. 
Notably, while these methods are online, they predominantly address setups with a single robot with multiple onboard sensors, rather than a truly distributed, multi-robot system.

In the field of sensor networks, CaliBree~\cite{Miluzzo:etal:DCSS2008} performs fully distributed sensor calibration by measuring disagreement between uncalibrated and calibrated sensors upon rendezvous event. 
This method, however, is only limited to calibration and does not address the localisation of the devices.
In~\cite{Devarajan:etal:IEEE2008}, GBP is used in a distributed fashion for intrinsic calibration and refinement of the camera poses. 
The method solves structure-from-motion, where multiple cameras are stationary; hence, it is not applicable to online robotic applications with a moving onboard sensor. 
Non-parametric belief propagation is used in~\cite{Ihler:etal:SensorNetwork2004} to perform calibration and localisation of sensors. The method is sampling-based; hence less efficient than GBP and assumes that the sensors are stationally.
LaSLAT~\cite{Taylor:etal:IPSN2006} performs localisation and calibration of the sensors together with tracking of a target. 
In LaSLAT, the sensor poses are assumed to be static and are not suitable for localising multiple moving robots, which is the problem we address in this work.

\subsection{Multi-Robot Localisation}
Many recent advancements in multi-robot localisation leverage the advancements in distributed pose-graph optimisation (PGO).
For example, \cite{Choudhary:etal:IJRR2017} uses chordal-relaxation to make the underlying PGO problem linear, and solves them using a Gauss-Seidel solver.
Semidefinite Programming (SDP) relaxation together with Riemannian block coordinate descent is used in \cite{Tian:etal:RAL2020,Tian:etal:TRO2021} which enables verification of the correctness of the estimates, and is decentralised and asynchronous. A distributed SLAM system~\cite{Tian:etal:TRO2022} is built upon these approaches, demonstrating their practicality.
However, the above approaches require a full relative transformation between the robots and can only handle isotropic covariance. This could be limiting, for example, if the inter-robot observations are all range-bearing, as we will explore in this paper.
More recently, methods use range measurements~\cite{Papalia:etal:arxiv2023} or bearing measurements~\cite{Wang:etal:RAL2022} and show that it is possible to obtain certifiably optimal solutions again via SDP relaxation.
While all PGO-based methods achieve good localisation accuracy, the formulation is often tailored and 
these methods do not generalise to other problem instances.

More general methods for multi-robot localisation such as DDF-SAM~\cite{Cunningham:etal:IROS2010}, and DDF-SAM2~\cite{Cunningham:etal:ICRA2013} operate on factor graphs. 
They rely on Gaussian elimination and require robots to exchange Gaussian marginals about shared variables.
The communication; however, increases quadratically with the number of shared variables. 
Alternating Direction Method of Multipliers (ADMM) has been employed for distributed SLAM~\cite{McGann:etal:ARXIV2023} or to efficiently share map points for distributed bundle adjustment~\cite{Baenninger:etal:ICRA2023}. However, unlike elimination-based approaches only the point-estimates are recovered.

Our recent work, Robot Web~\cite{Murai:etal:TRO:2023} uses GBP to perform localisation. 
The proposed approach operates on general factor graphs; the underlying GBP algorithm is asynchronous and decentralised, thus the system is scalable to many robots.

\section{PRELIMINARIES}
\subsection{Factor Graphs}
A factor graph is a type of bipartite graph $G=(X, F, E)$ consisting of variable nodes $X = \{ \vecx_i \}_{i=1:N_v}$ connected by edges $E$ to factor nodes $F = \{f_s\}_{s=1:N_f}$.
A factor graph represents the factorisation of the joint distribution: $p(X) \propto \prod_{s=1}^{N_f} f_s(X_s)$, where $X_s = n(f_s)$. Here, $n(v)$ is the set of neighbouring nodes connected via edge to the node $v$.

\subsection{Belief Propagation}
Belief Propagation (BP)~\cite{Pearl:AAAI:1982} is an algorithm used to infer a marginal distribution of $p(\vecx_i)$ for each variable $\vecx_i \in X$ of a joint distribution $p(X)$ by passing messages among neighbouring nodes.
As such operation is local, the algorithm is highly parallelisable and suitable for distributed inference. BP has promising applications in robotics, including multi-hypothesis and non-Gaussian inference~\cite{Dehann:etal:IROS2016}.

In BP, at each iteration $t$, factor $f_s$ sends a message $m^{t}_{f_s \rightarrow \vecx_i}$ to all variables $\vecx_i\in n(f_s)$, and variable $\vecx_i$ sends back a message $m^{t}_{\vecx_i \rightarrow f_s}$ to all factors $f_s \in n(\vecx_i)$.
At a variable $\vecx_i$, the product of all the incoming messages forms the belief $b^{t}(\vecx_i)$:
\beq
b^{t}(\vecx_i) = \prod_{f_s \in F_i}{m^{t-1}_{f_s \rightarrow \vecx_i}}(\vecx_i)~, 
\eeq
where $F_i = n(\vecx_i)$.
The outgoing message to $f_s$ from $\vecx_i$ is simply the product of all other incoming messages from $F_i$:
\beq
m^{t}_{\vecx_j \rightarrow f_s}(\vecx_j) = \frac{b^{t}(\vecx_i)}{m^{t-1}_{f_s \rightarrow \vecx_i}(\vecx_i)}~.
\eeq
The message from a factor to a variable message is: 
\beq
m^{t}_{f_s \rightarrow \vecx_i}(\vecx_i) = \sum_{X_s \setminus \vecx_i} f_s(X_s) \prod_{\vecx_j \in X_s \setminus \vecx_i} m^{t}_{\vecx_j \rightarrow f_s}(\vecx_j)~,
\eeq
where $X_s = n(f_s)$. A factor takes a product of all the incoming messages together with its potential and marginalises out all the other adjacent variables $X_s \setminus \vecx_i$ to compute the outgoing message to $\vecx_i$.

\subsection{Gaussian Belief Propagation}
GBP~\cite{Weiss:Freeman:NIPS2000} is a subset of BP where all the factors and, hence, the joint posterior distribution are Gaussians.
Note that GBP has no theoretical convergence guarantees when applied to graphs with cycles and yet, it has shown robust performance across many different tasks
~\cite{Patwardhan:etal:RAL2023,Ortiz:etal:CVPR2020,Scona:etal:RAL2022}.

We use canonical representation of Gaussian: $\cN^{-1}(\vecx; \veceta, \matLambda) \propto \exp(-\frac{1}{2} \vecx^{\top} \matLambda x + \veceta^{\top} \vecx)$, where we define information vector $\veceta = \matLambda \vecmu$ and information matrix $\matLambda = \matSigma^{-1}$.
Measurements  $Z = \{\bar{\vecz}\}_{m=1:N_z}$ are modelled as $\bar{\vecz}_m = h_m(X_m) + \epsilon$, where $h_m(\cdot)$ is a measurement prediction function, and $\epsilon$ is a zero-mean Gaussian $\epsilon \sim \cN(0, \matSigma_m)$.
Assuming independence of the observations, the likelihood of the observations is:
\bea
&l(X ; Z)  = \prod_{m}^{N_z}l_m(X_m ; \bar{\vecz}_m) \nonumber \\
&\propto \prod_{m}^{N_z}\exp(-\frac{1}{2}\|\bar{\vecz}_m - h_m(X_m)\|_{\Sigma_M}^{2})~,
\eea
where the likelihood $l(X ; Z) \propto p(Z | X)$. This notation is used to clarify that likelihood is a function of $X$, with $Z$ as its parameter~\cite{Dellaert:Kaess:2017}.
We are interested in estimating the configuration which maximises the Maximum A Posteriori (MAP) estimate:
$X^{MAP} = \argmax_X l(X;Z) P(X)$.
GBP achieves this via marginal inference by computing the marginal posterior $P(\vecx_i | Z) = \cN(\vecmu_i, \matSigma_i)$ where $\vecmu_i = \vecx_i^{MAP}$ for all $\vecx_i \in X$.
In the factor graph representation of the posterior $P(X|Z)$, each $\vecx_i$ is a variable node, and each $l_m$ is a factor node connected via edge to nodes $X_m$.
The reader is referred to~\cite{Ortiz:etal:arxiv2021,Davison:Ortiz:arxiv2019} for more details on GBP.

\subsection{Gaussian Belief Propagation with Lie Groups}
Given a non-linear measurement prediction function $h_m$ it's first-order Taylor approximation is: $h_m(\vecx) \approx h_m(\vecx_0) + \matJ_m h_m(\vecx - \vecx_0)$, $\matJ_m = \frac{dh_m}{d\vecx}|_{\vecx=\vecx_0}.$
We retrieve the likelihood of the descent direction $\Delta \vecx = \vecx - \vecx_0$ that minimises the local residual as in \cite{Murai:etal:TRO:2023}:
\bea
\label{eq.nonlinear_likelihood}
&l_m(\Delta X_m; \bar{\vecz}_m) = \nonumber \\ 
&\cN^{-1}(\Delta X_m; \matJ_m^T\matLambda_m(\bar{\vecz}_m-h_m(\vecx_0)), \matJ_m^T\matLambda_m\matJ_m)~.
\eea
For clarity,  the $\Delta$ is dropped from the notation of the likelihood for the remaining of the paper.
Energy of the factor is defined as: $E_m = (\bar{\vecz}_m - h_m(x_0))^{\top} \matLambda_m (\bar{\vecz}_m - h_m(x_0))$.

We further apply GBP to variables on the Lie Group as in \cite{Murai:etal:TRO:2023}, which rely on the following expressions:
\bea
\cY &= \cX \oplus {}^\cX\tau \triangleq \cX \circ \Exp({}^\cX\tau) \in \cM~, \\
{}^\cX\tau &= \cY \ominus \cX \triangleq \Log(\cX^{-1} \circ \cY) \in T_\cX\cM~,
\eea
where $\cX, \cY$ are points on the Lie Group $\cM$, and ${}^\cX\tau$ is a tangent vector in Lie algebra $T_\cX\cM$ defined locally around a point $\cX$. Functions Exp, Log are the exponential, logarithmic mapping from $\cM \rightarrow T_\cX\cM$ and $T_\cX\cM  \rightarrow  \cM$ respectively, allowing us to move back and forth between Lie Group and Lie algebra. Composition operation $\circ$ uses the group property and imposes that the composition of the elements remains in the Lie Group. To simplify the notation, we will drop $\circ$ when composing two transformations.

\section{DISTRIBUTED LOCALISATION AND EXTRINSIC CALIBRATION }

\begin{figure}[!tbp]
  \center
  \includegraphics[width=\linewidth]{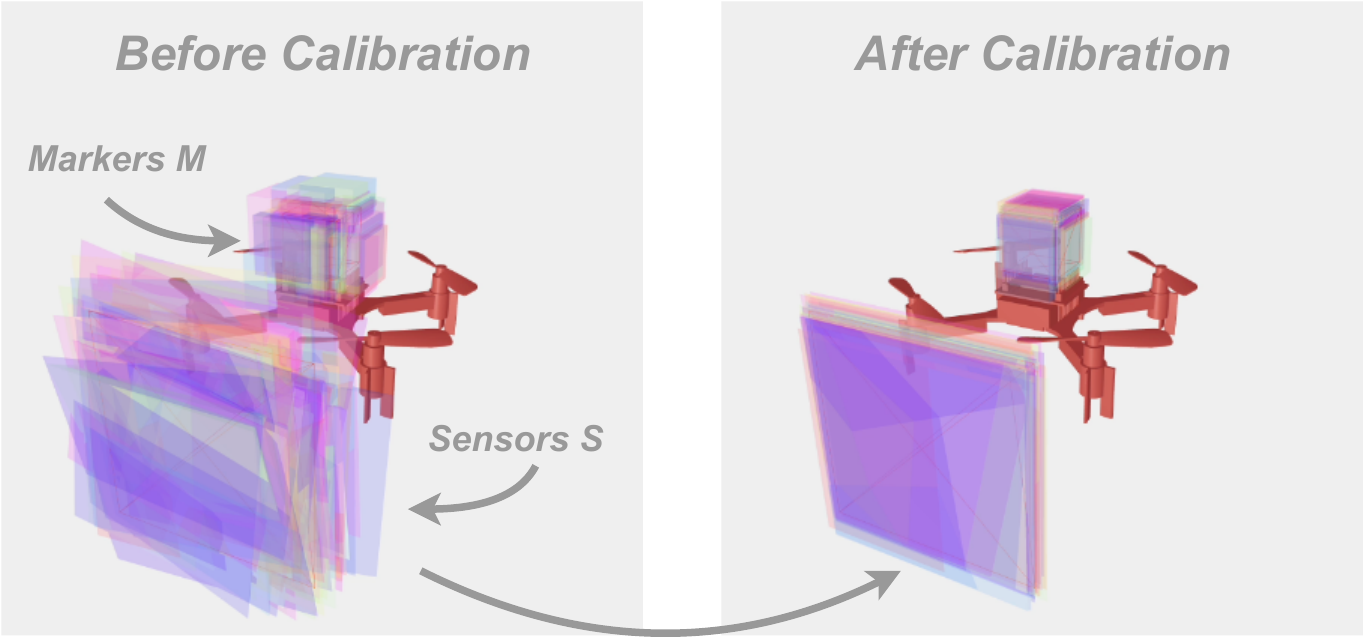}
  \caption{
  Example of calibration of the extrinsic of the sensors' pose and markers' position using the proposed method, 
  where we artificially set the ground-truth extrinsics to be the same for visual clarity.
  We overlay the calibration estimates of 64 robots from randomly initialised states (left), and visualise the estimated extrinsics after the calibration (right).
  }\label{fig:before-after-calibration}
    \vspace{2mm} \hrule
    \vspace{-2em}
    \vspace{2mm}
\end{figure}

\subsection{Gaussian Belief Propagation for Distributed Inference}
Let $G$ be the global factor graph which we want to perform inference over.
In distributed GBP, robots own a factor graph $G^\omega$ each, and their union is $G  = \bigcup_{\omega\in\Omega} G^\omega$.
Each robot {\em has ownership} of its pose variables and the factors corresponding to the observations it made, i.e. the nodes that their local graph $G^\omega$ consists of.
This is important as the global factor graph $G$ is partitioned amongst the robots $\Omega$; hence, the marginal estimates obtained by solving the distributed problem using GBP are {\em exactly} the same as the marginal estimates of the global problem obtained via centralised GBP under an assumption of perfect communication.

Distributed inference is achieved by each robot $\alpha\in\Omega$ performing GBP message passing on their local graph $G^\alpha$. 
Along the edges of a factor $f^{\alpha \beta}$ owned by $\alpha$, a factor-to-variable message ${m_{f^{\alpha \beta} \rightarrow x^\beta}}(x^\beta)$ is sent from $\alpha$ to $\beta$   via inter-robot communication. Similarly, $\beta$ sends back to $\alpha$ variable-to-factor message, ${m_{x^\beta \rightarrow f^{\alpha \beta}}}(x^\beta)$. 
A factor-to-variable message is a Gaussian distribution $\cN(\mu_{f\rightarrow x}, \Lambda_{f\rightarrow x}^{-1})$, and a variable-to-factor message is a linearlisation point $\xbar$ and a Gaussian distribution  $\cN(\mu_{x\rightarrow f}, \Lambda_{x\rightarrow f}^{-1})$.

\subsection{Problem Formulation}\label{sec:problem-formulation}
This section details how multi-robot localisation and the extrinsic calibration of their onboard sensors and observable markers are simultaneously and distributedly performed.
In the considered setting, each robot is equipped with a range-bearing sensor $S$ that observes the other robot's onboard marker $M$.
While the addressed setup is representative of realistic constraints of many robotic applications, here we describe the formulation in general terms, so that it can easily be extended to support more information-rich measurements such as direct relative transformations (e.g. using visual sensors and fiducial markers).

The relative transformation from the base of the robot $B$ to its sensor $S$ is $\matT_{BS} \in \SE3$, where the notation $\matT_{BS}$ represents the pose of $S$ in the coordinate frame of $B$.
Similarly, $\vect_{BM} \in \RR^3$ represents the marker position $M$ relative to $B$. 
Since only range-bearing sensors are used in the current setup, the orientation of the marker is not observable and thus not included in this specific problem definition.
When the sensor $S^\alpha$ in robot $\alpha$ observes the marker $M^\beta$ in robot $\beta$, a relative measurement $\bar{\vecz}^{\alpha\beta}_{SM}$ is generated. 
The initial estimates of $\matT_{BS}$ and $\vect_{BM}$ are expected to be noisy due to inaccurate calibration. 
Our work optimises over the extrinsic calibration using the observations $\bar{\vecz}^{\alpha\beta}_{SM}$ robots accumulate over time as depicted in Fig.~\ref{fig:before-after-calibration}.

Let $\Omega = \{ \alpha, \beta, \gamma, \ldots \}$ be the set of robots, $T$ be the number of considered time-steps, and $\matT^{\omega}_{WBt}$ denote the pose of the base of the robot $\omega$ at time $t$ in the world coordinate.
To perform marginal inference over all, $\matT^{\omega}_{BS}, \vect^{\omega}_{BM}, \matT^{\omega}_{WBt}, \forall \omega \in \Omega, \forall t \in \{1, \ldots T \} $, we  consider the following factors.

\subsubsection{Range Bearing Sensor}
In our setup, robots can observe the other robots using range-bearing sensors. 
We use spherical coordinate $({r}, {\theta}, {\phi})$, i.e. radial distance, azimuthal angle, and elevation angle respectively. 
All angles are parameterised using $\SO2$, hence; the range bearing measurement is $\bar{\vecz}^{\alpha\beta}_{SM} \in \langle \RR, \SO2, \SO2 \rangle$, a composite manifold. 

The range bearing factor relating the sensor $\matT^\alpha_{WSt}$ and the marker $\vect^\beta_{WMt}$ at time $t$ is:
\bea
&l_s(\matT^\alpha_{WSt}, \vect^\beta_{WMt}; \bar{\vecz}^{\alpha\beta}_{SM}) \propto \nonumber \\
&\exp(-\frac{1}{2} \| \bar{\vecz}^{\alpha\beta}_{SM}  \dminus h_s(\matT^\alpha_{WSt}, \vect^\beta_{WMt}) \|_{\Sigma_s}^{2} )~,
\label{eq.rangebearing_likelihood}
\eea
where we use the notation  $\dminus$ from~\cite{Sola:etal:arxiv2018}, an operation on the composite manifold ($\ominus$ operation is applied to each block of composites separately), and $h_s$ is the function that predicts range bearing measurement between $\matT^\alpha_{WS}$ and $\vect^\beta_{WM}$.

\subsubsection{Robot Odometry}
We assume that odometry measurements $\bar{\matT}_{B_{t-1} B_t} \in \SE3$  (e.g. IMU/wheel odometry) are made available to each robot. 
An odometry factor penalises the deviation between observation $\bar{\matT}_{B_{t-1} B_t} $ and the two estimated consecutive poses $\matT_{WB_{t-1}},  \matT_{WB_t}\in \SE3$:
\bea
&l_o(\matT_{WB_{t-1}},  \matT_{WB_t}; \bar{\matT}_{B_{t-1} B_t}) \propto \nonumber \\ 
&\exp(-\frac{1}{2} \| \bar{\matT}_{B_{t-1} B_t} \ominus (\matT_{WB_{t-1}}^{-1}  \matT_{WB_t}) \|_{\Sigma_o}^{2} )~.
\label{eq.odometry_likelihood}
\eea
This assumes that the odometry measurement is measured in the base frame $B$.
This property can be enforced by choosing a suitable base frame given prior information about the wheel/IMU position.
However, if the odometry is provided via the sensor $S$ (i.e. visual odometry), we can replace the transformations in the base frame $B$ with a transformation in the sensor from $S$.

\begin{figure}[!tbp]
  \center
  \includegraphics[width=\linewidth]{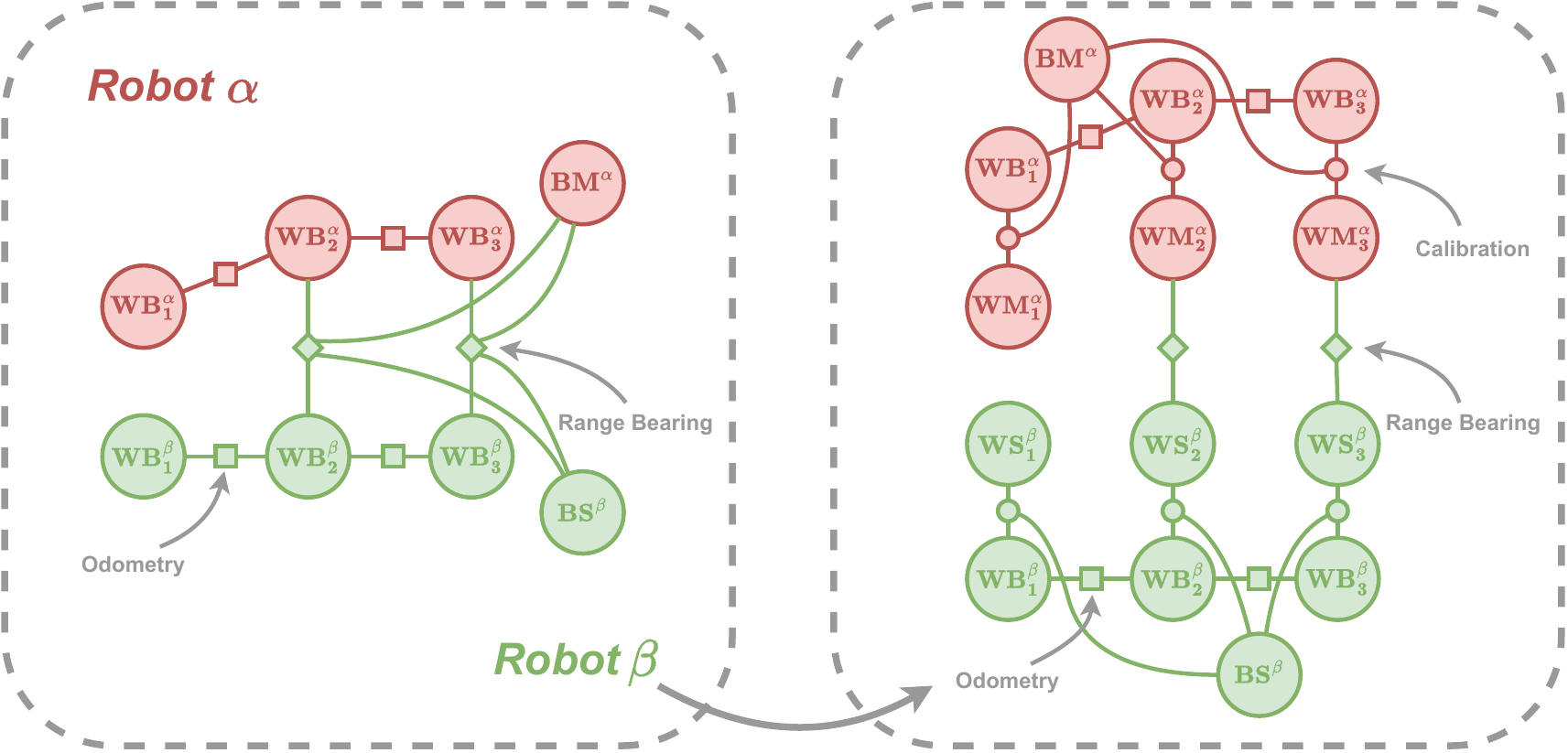}
  \caption{
Reducing the inter-robot communication by restructuring the factor graph. We refer to $\matT_{WB}$ as $\matW\matB$ for clarity. \textbf{Left:} The inter-robot factor (range bearing) depends on four variables: the poses of the robots, the marker $M$, and the sensor $S$ pose with respect to the robot base $B$.  
\textbf{Right} We introduce the marker and sensor variable in the world coordinate frame using Eq.~\eqref{eq.calibration_likelihood}. 
While the total number of variables increases, the inter-robot factors depend on fewer variables, thus reducing the communication requirements.
}\label{fig:calibration-factor-graph}
    \vspace{2mm} \hrule
    \vspace{-2em}
    \vspace{2mm}
\end{figure}

\subsubsection{Calibration Factor}
In Eq.~\eqref{eq.rangebearing_likelihood}, we have used $\matT_{WSt}, \vect_{WMt}$ position of sensor $S$ and marker $M$ in the world coordinate frame $W$ at time $t$.
A simple solution to obtain the position in the world coordinate is to use all $\matT^\alpha_{WBt}, \matT^\alpha_{BS}, \matT^\beta_{WBt}, \vect^\beta_{BM}$ inside the likelihood function, as shown in the left row of Fig.~\ref{fig:calibration-factor-graph}.
However, this has a clear disadvantage: robots must communicate both the calibration estimate and the pose estimate with each other. This not only doubles the inter-robot communication effort but also exposes internal states (i.e. sensor calibration) that do not need to be revealed to other robots. Furthermore, it creates small cycles which often leads to overconfidence~\cite{Weiss:Freeman:NIPS2000}.
Hence, this motivates the redesign of a factor to only share the pose estimate of the sensor and the marker between robots.

The objective of the calibration is to find a transformation $\matT_{BS}\in \SE3$ such that:
$\matT_{WS} =\matT_{WB} \matT_{BS}$.
This relationship as a likelihood is defined as:
\bea
&l_c(\matT_{WS}, \matT_{WB}, \matT_{BS}) \propto \nonumber \\ 
&\exp(-\frac{1}{2} \| \Log (\matT_{WS}^{-1}  \matT_{WB} \matT_{BS}) \|_{\Sigma_c}^{2} )~.
\label{eq.calibration_likelihood}
\eea
The likelihood for calibration of the marker can be derived in a similar way.
This allows us to create a factor graph as illustrated in the right row of Fig.~\ref{fig:calibration-factor-graph}, where only $\matT_{WSt}, \vect_{WMt}$ is connected between the robots. 
While this formulation increases the total number of variables in the factor graph, fewer variables are connected to the inter-robot factor, thus reducing the data transfer between the robots. 

\subsection{Adaptive Regulariser on the Factor}
Due to the nature of \SE3, the objective function which we are minimising is non-linear and non-convex; challenging for any iterative optimisers, but especially for local ones such as GBP with no access to the global objective function.
In our case, the Lie group extension of GBP~\cite{Murai:etal:TRO:2023} was insufficient to consistently reach convergence.
Hence, here, we introduce an adaptive regularisation term in GBP to assist convergence.

For each of the factors $f_m$, with a likelihood as defined in Eq.~\eqref{eq.nonlinear_likelihood}, we add a zero-mean prior $\cN^{-1}(0, \lambda_m \matI)$. The term $\lambda^t_m$ is local to the factor and is updated adaptively based on the difference between the current local factor energy $E_m^t$ and last iterations $E_m^{t-1}$:
\beq
\lambda_m^t = 
    \begin{cases}
     \lambda_m^{t-1} \cdot \lambda_{\uparrow} & E^t_m - E^{t-1}_m > \epsilon_\lambda \\
     \lambda_m^{t-1} / \lambda_{\downarrow} &\text{otherwise} \\
    \end{cases}~,
\eeq
where a threshold $\epsilon_\lambda$ is required to avoid the weighting from increasing when the factors' energy stops changing significantly near convergence, and $\lambda_\uparrow, \lambda_\downarrow$ are the increase, decrease factor respectively.
Intuitively, at the beginning when far from optima, the adaptive regulariser encourages small descent steps.
As the factors become more confident about its approximation of the curvature (i.e. made multiple successive descents), larger descent steps are performed.
While the principle of this approach is the same as Levenberg-Marquardt, fundamental 
this weighting scheme is computed and applied purely locally, and the step is always taken even if the local energy increases.
This way, no synchronisation or communication is required when applied distributedly.

Assuming that the objective function is strictly convex, the addition of the adaptive regualisation term will not change the optimal solution.
As the GBP converges, $\lim_{\lambda_m \rightarrow 0} l_s(X_s;  \bar{\vecz}_s) \cN^{-1}(0, \lambda_m\matI) =  l_s(X_s; \bar{\vecz}_s)$, and $\lambda_m \rightarrow 0$ as the energy decreases or reaches local convergence.

\section{EVALUATION}
\begin{table}[!t]
\caption{
Accuracy of the proposed method (`Ours') and the global, centralised NLLS LM solver (`LM') at convergence as a function of the number of robots $N$ and the enabling of autocalibration.
Results include the RMSE ATE and ARE  of the robot poses of their bases in the world frame $\matT_{WB}$, the extrinsic calibration of their sensor $\matT_{BS}$  and marker $\vect_{BM}$ (only translation) where applicable.
\label{table:comparison_against_batch_solver}
}

\centering
\setlength\tabcolsep{2.5pt}
\begin{tabular}{cc|cc|cccccc}
\hline
N & $\matT$ & \multicolumn{2}{c|}{Initial} & \multicolumn{2}{c}{LM  w/ Calib.} & \multicolumn{2}{c}{Ours w/ Calib.} & \multicolumn{2}{c}{Ours w/o Calib.}  \\

  & & [m] & [deg] & [m] & [deg] & [m] & [deg] & [m] & [deg] \\
\hline

16 & $\matT_{WB}$ & 0.432 & 7.422 & 0.065 & 1.858 & 0.084 & 1.970 & 0.093 & 2.313 \\
   & $\matT_{BS}$ & 0.080 & 8.852 & 0.023 & 1.156 & 0.027 & 1.268 & --   & --   \\
   & $\vect_{BM}$ & 0.085 & --   & 0.020 & --   & 0.022 & --   & --   & --   \\

\hline
32 & $\matT_{WB}$  & 0.434 & 7.471& 0.051 & 1.742& 0.062 & 1.811 & 0.075 & 2.138\\
   & $\matT_{BS}$  & 0.082 & 8.856& 0.021 & 1.035& 0.025 & 1.251 & --   & --  \\
   & $\vect_{BM}$  & 0.087 & --  & 0.019 & --  & 0.022 & --   & --   & --  \\

\hline
64 & $\matT_{WB}$ & 0.436 & 7.402& 0.043 & 1.684& 0.054 & 1.761& 0.066 & 2.082 \\
   & $\matT_{BS}$ & 0.083 & 8.810& 0.020 & 0.969& 0.025 & 1.214& --   & --   \\
   & $\vect_{BM}$ & 0.088 & --  & 0.018 & --  & 0.020 & --  & --   & --   \\

\hline
128 & $\matT_{WB}$  & 0.434 & 7.385 & 0.039 & 1.646 & 0.049 & 1.732& 0.060 & 2.041  \\
    & $\matT_{BS}$  & 0.085 & 8.740 & 0.018 & 0.969 & 0.022 & 1.202&  --  &  --   \\
    & $\vect_{BM}$  & 0.087 & --   & 0.017 & --   & 0.020 &  -- &  --  &  --   \\

\hline%

\end{tabular}
\vspace{-2em}
\vspace{2mm}
\end{table}

\begin{figure}
  \centering
  \begin{subfigure}[b]{0.49\linewidth}
    \includegraphics[height=4cm]{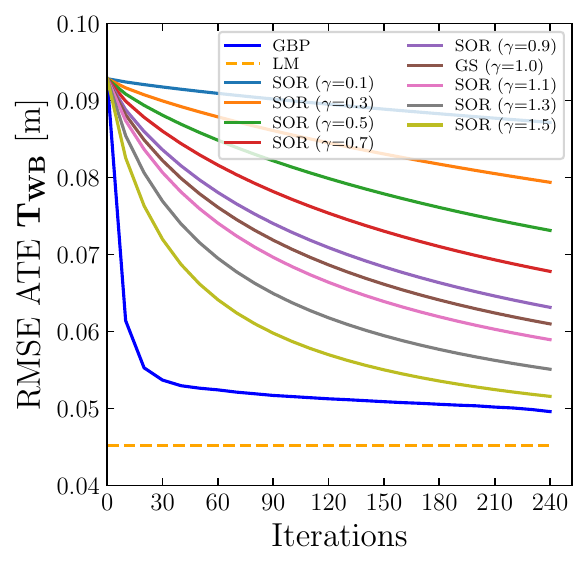}
  \end{subfigure}
  \hfill
  \begin{subfigure}[b]{0.49\linewidth}
    \includegraphics[height=4cm]{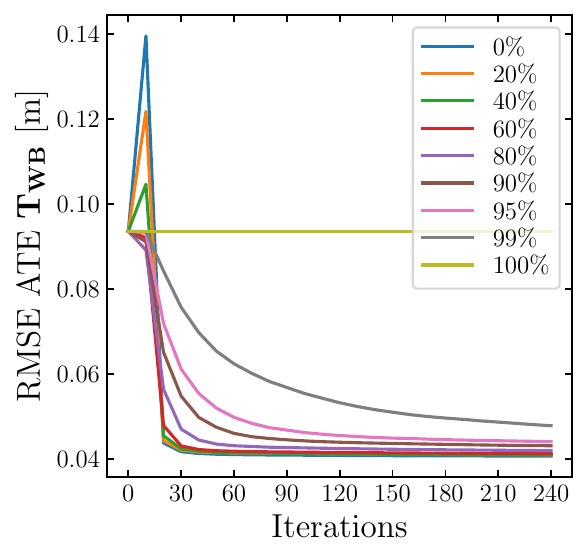}
  \end{subfigure}

  \begin{subfigure}[b]{0.49\linewidth}
    \includegraphics[height=4cm]{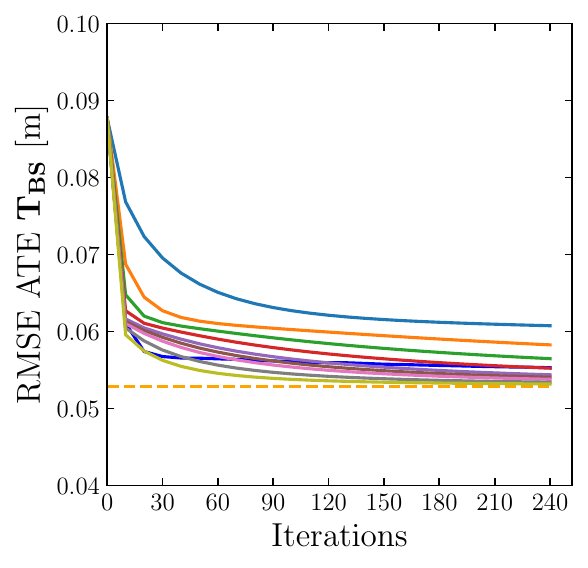}
  \end{subfigure}
  \hfill
  \begin{subfigure}[b]{0.49\linewidth}
    \includegraphics[height=4cm]{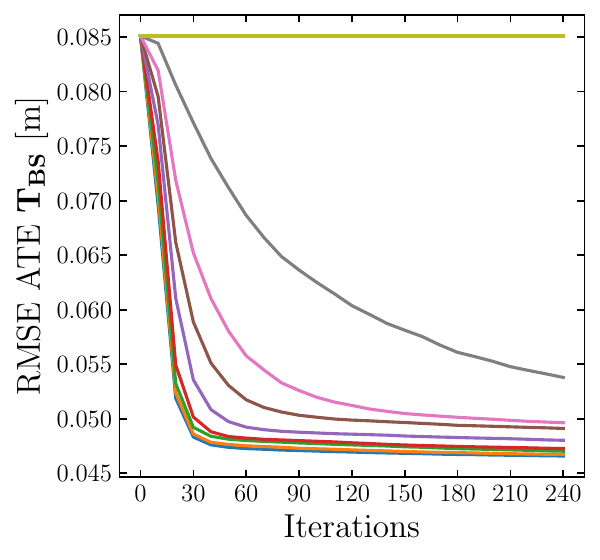}
  \end{subfigure}
  
  \begin{subfigure}[b]{0.49\linewidth}
    \includegraphics[height=4cm]{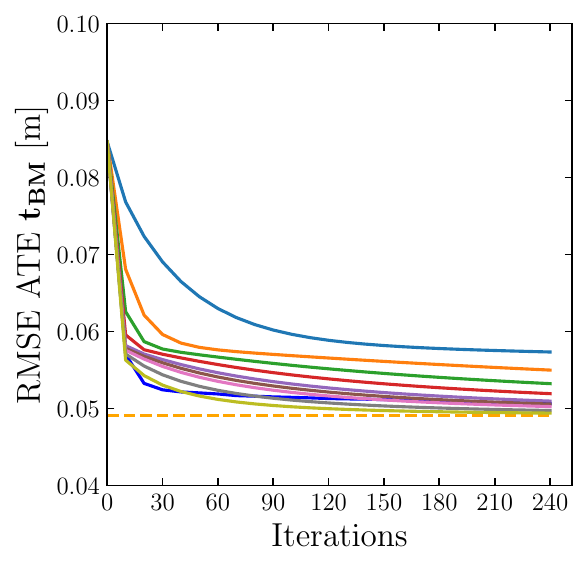}
  \end{subfigure}
  \hfill
  \begin{subfigure}[b]{0.49\linewidth}
    \includegraphics[height=4cm]{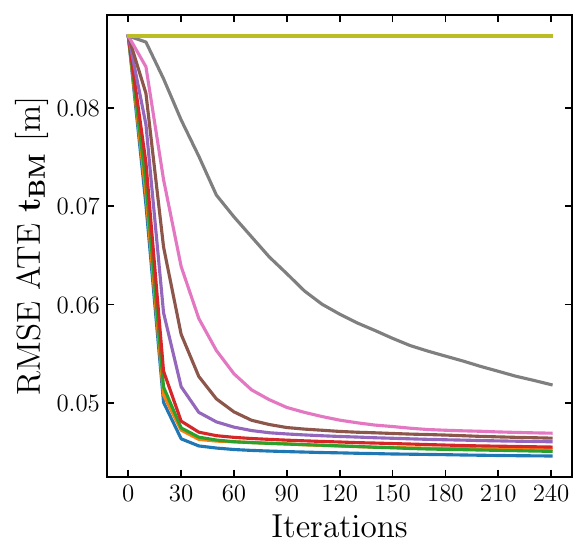}
  \end{subfigure}
  
  \caption{
    From top to bottom: RMSE ATE for $\matT_{WB}$, $\matT_{BS}$, $\vect_{BM}$. RMSE ARE omitted as it follows the same trend.
  \textbf{Left:} Comparison of different distributed alternatives (Final RMSE ATE of global, non-distributed LM shown for reference). 
  \textbf{Right:}  Analysis of robustness regarding communication failures by randomly dropping a percentage of the inter-robot GBP messages in each iteration. 100\% indicates that all inter-robot messages are dropped, preventing co-localisation.}
  \label{fig:dsolver_dropout}
    \vspace{2mm} \hrule
    \vspace{-2em}
    \vspace{2mm}
\end{figure}

\begin{figure}
  \centering
  \begin{subfigure}[b]{0.49\linewidth}
    \includegraphics[height=4cm]{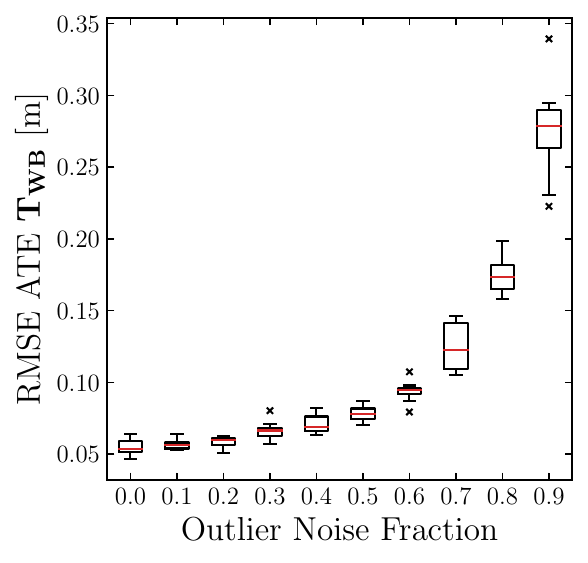}
  \end{subfigure}
  \hfill
  \begin{subfigure}[b]{0.49\linewidth}
    \includegraphics[height=4cm]{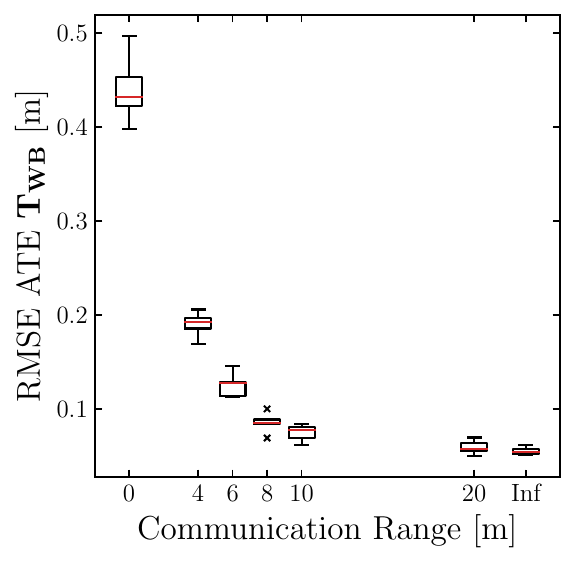}
  \end{subfigure}

  \begin{subfigure}[b]{0.49\linewidth}
    \includegraphics[height=4cm]{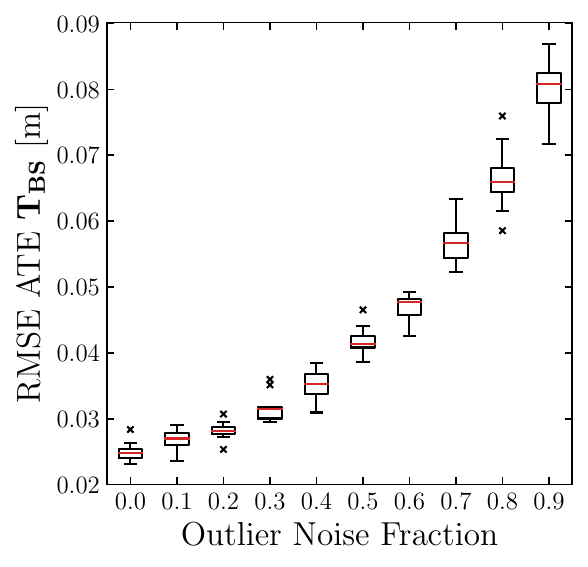}
  \end{subfigure}
  \hfill
  \begin{subfigure}[b]{0.49\linewidth}
    \includegraphics[height=4cm]{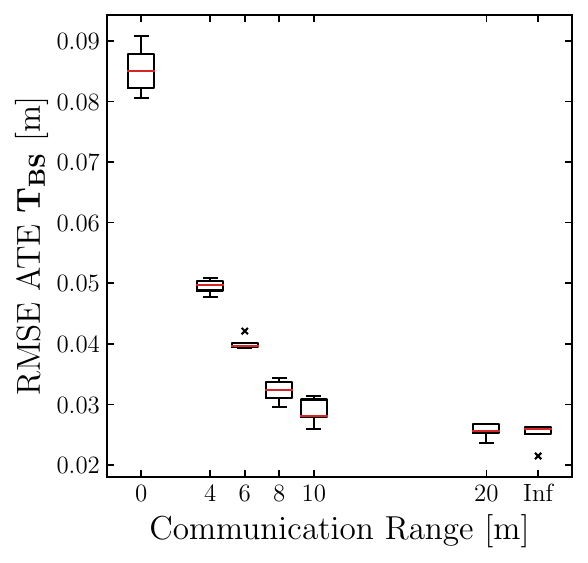}
  \end{subfigure}
  
  \begin{subfigure}[b]{0.49\linewidth}
    \includegraphics[height=4cm]{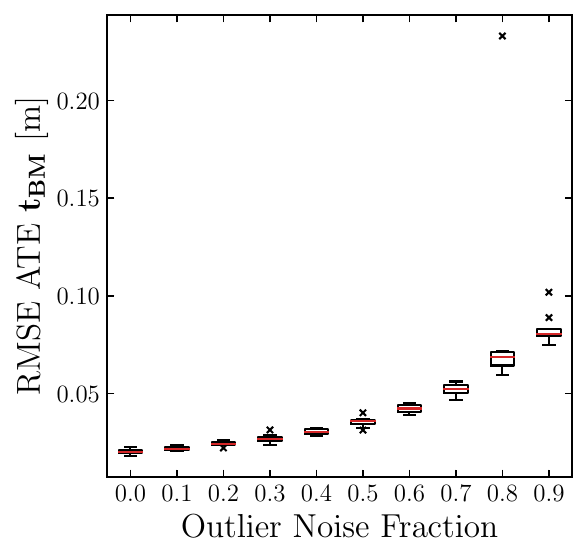}
  \end{subfigure}
  \hfill
  \begin{subfigure}[b]{0.49\linewidth}
    \includegraphics[height=4cm]{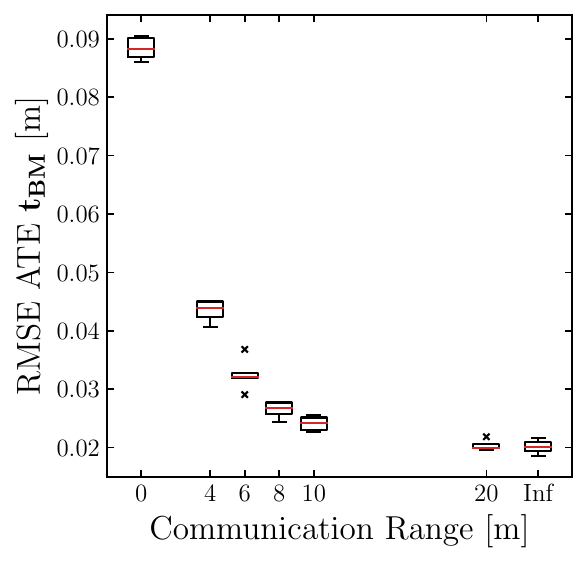}
  \end{subfigure}
  
  \caption{
  From top to bottom: RMSE ATE for $\matT_{WB}$, $\matT_{BS}$, $\vect_{BM}$. RMSE ARE omitted as it follows the same trend.
  \textbf{Left:} Effect of increasing the fraction of outlier noise. Non-Gaussian noise is added to the inter-robot sensor measurement to simulate outliers.
  \textbf{Right:}  Impact on the overall accuracy when robots are limited to only communicating with peers within the specified range.}
  \label{fig:oulier_comm_range}
\vspace{2mm} \hrule
\vspace{-2em}
\vspace{2mm}
\end{figure}

We mainly evaluate our approach in a simulated environment with a vast number of robots, as obtaining the ground-truth extrinsic calibration and robot poses in the real-world for such experiments would be extremely challenging. As a verification of the applicability of our method to the real-world, we evaluate using UTIAS MR.CLAM dataset~\cite{Leung:etal:IJRR2011}.

To simulate sensor noise, observations are corrupted by applying zero-mean Gaussian noise.
Odometry measurements are corrupted with noise with $\sigma_B^t$, a standard deviation of 0.01 meter per meter travelled for the translation, and with $\sigma_B^R$, a standard deviation of 1 degree per 90 degrees rotated for the rotation.
Inter-robot measurements are corrupted in their range and bearing readings with a standard deviation of $\sigma_s$: (0.05m, $5\deg$).
3D range bearing measurements $(r, \theta, \phi)$ are restricted to the three closest observable robots, to imitate realistic and limited
inter-robot observability. 
We further restrict the range-sensing to be limited to $|\theta| < 60\deg$ and $|\phi| < 60\deg$, to simulate the field of view limitations of, for instance, visual sensors.
Robots are randomly initialised in translation and orientation within a $20m \times 20m \times 20m$ space. 
We assume that the initial pose of the robots is known to a certain degree, within a noisy initial guess with standard deviation 0.01m, $1\deg$ respectively for the translation and the rotation.
We simulate the robots' motion by drawing random samples from a uniform distribution, $\cU(0, 1)m$ for translational motion and $\cU(-\pi, \pi)\deg$ for rotational motion, across all three dimensions.
Finally, the initial calibration of sensor $S$ and marker $M$ also deviate from the ground-truth with a standard deviation of the translational part of extrinsic of sensor $\sigma_S^t$ and marker $\sigma_M^t$
set to $0.05m$, and the standard deviation of the rotation part of the sensor frame $\sigma_S^R = 5\deg$.

To enhance the stability of GBP, for the adaptive regularisation, we use the default parameter of $\lambda_m = 10, \lambda_\downarrow = 9, \lambda_\uparrow = 11$, $\epsilon_\lambda = 10^{-4}$. While not sensitive to the choice of parameters, we found GBP to diverge in many cases without adaptive regularisation.
Additionally, 30\% of both internal and external GBP messages are randomly dropped, as an empirical heuristics to improve the convergence of the system~\cite{Ortiz:etal:ICRA2022}.
Unless specified otherwise, for robustness against outlying measurements, we dynamically scale the information matrix of the range-bearing sensor factor using a DCS robust kernel~\cite{Agarwal:etal:ICRA2012}:
$
s_m = \min \left(1, \frac{2\Phi}{\Phi + E_m}\right),
$
with $\Phi=10$. For each range-bearing measurement, the information matrix $\matLambda_m$ is scaled by $s_m^2$. 

In all the presented experiments, unless specified otherwise, we consider $N=64$ robots that randomly execute $50$ motions, incrementally growing the underlying factor graph (see Section~\ref{sec:problem-formulation}) and performing $30$ GBP message-passing iterations after each of these motions.
The experimental results aggregate information from a total of 10 randomised runs, where we often
report the average Root Mean Squared Error (RMSE) of Absolute Trajectory Error (ATE) and Absolute Rotation Error (ARE) to measure the accuracy of the system as described in~\cite{Choudhary:etal:IJRR2017} for multi-robot setup. 

\subsection{Comparison with Centralised Factor-Graph Solvers}
Here we compare the proposed incremental GBP-based approach with a global Non-Linear Least Squares (NLLS) Levenberg-Marquardt (LM) solver (implemented in Theseus~\cite{Pineda:etal::NeurIPS2022}) that processes the full graph as a whole batch.
We evaluate how the accuracy of the overall system varies as a function of the number of robots $N$ and the effect resulting from enabling or disabling auto-calibration for the proposed method,  i.e. whether the initial noisy calibration is optimisable or remains fixed, respectively. The robust kernel is disabled for this experiment to simplify the comparison.
The accuracy of the different alternatives is compared in Table~\ref{table:comparison_against_batch_solver}.
Despite the proposed method being distributed and without any global, second-order perspective of the whole problem, experimental results show no significant differences with respect to the global LM solver at convergence.
As expected, the larger the number of robots $N$, the higher the accuracy of all methods as the underlying factor graph becomes denser and thus, more information-rich.
Observe that the proposed GBP-based approach is still able to profit from a denser graph despite including more cycles.
We additionally report the results of our method while considering that the noisy initial calibration is correct.
This yields obviously worse results than when we optimise the graph which considers the calibration parameters.

\subsection{Comparison with Distributed Factor-Graph Solvers}\label{sec:comp_batch}
We compare our method against other distributed solvers: block Gauss-Seidel (GS) and its relaxations block Successive Over-Relaxation (SOR)~\cite{Dimitri:2015}.
In GBP messages are exchanged in parallel and thus do not require coordinated updates.
We favour GS and SOR by counting each iteration as an ordered sweep, where robots sequentially exchange their updated state in a specific order. In this comparison, all the methods are provided with the whole graph to be optimised from the beginning instead of incrementally growing and solving the problem, with 16 robots making 10 random motions. We use the same relaxation parameter as reported in~\cite{Choudhary:etal:IJRR2017}. The robust kernel is disabled for this experiment to simplify the comparison.

Results are presented in the left column of Fig.~\ref{fig:dsolver_dropout}. GBP shows a faster convergence rate than GS and SOR in the number of iterations (the aforementioned global and centralised LM method is also shown for reference).
Presented results match prior comparisons between GBP and Successive Over-Relaxation in~\cite{Weiss:Freeman:NIPS2000}.
Note that GS and SOR produce marginally better extrinsic at by trading off a significantly worse body frame localisation.

\subsection{Robustness Analysis}
\subsubsection{Communication Failure and Asynchronicity}
We analyse the robustness of the system regarding potential communication failures, modelled by randomly dropping a percentage of the inter-robot GBP messages in each iteration, and present the results in the right column of Fig.~\ref{fig:dsolver_dropout}.
In this experiment, the whole graph is available from the beginning of the proposed algorithm with 16 robots and 50 random motions to clearly identify the convergence trends.
The behaviour of the system remains largely unaffected by communication failure up to around 80\%, communication failure. ATE of $\matT_{WB}$ initially increases as poses are initially uncertain and are down-weighted by the robust kernel. However, within a few iterations, the poses are correctly optimised, reducing the ATE.
Even at an extremely high communication failure rate, the RMSE ATE still gradually decreases as we perform more iterations and thus, more rounds of communication. 
The experiment further demonstrates the asynchronicity of our approach, where the message order does not significantly impact the overall performance.
This is a crucial property required for real-world deployment, where the communication channel is potentially unreliable, especially at scale.

\subsubsection{Robustness to Outlier Measurements}
As real-world inter-robot sensing is often challenging (e.g. misidentification, sensor failure), we investigate the robustness of the system to extreme, non-Gaussian outlier measurements following a uniform noise.  
As presented in the left row of Fig.~\ref{fig:oulier_comm_range}, results indicate that, while performance is reasonably impacted as the fraction of outliers increases, the system remains stable even for an extremely high percentage of non-Gaussian outliers. 
Even at 40\%, we observe a relatively small increase in the error compared to no outlying noise, demonstrating the robustness of our approach.

\subsubsection{Communication Range}
To mimic realistic, real-world conditions, we further limit the communication radius of the robots to 4, 6, 8, 10 and 20m and report the results on the right row of Fig.~\ref{fig:oulier_comm_range}, including also no communication (0m) and infinite communication range (`Inf') for completeness.
While a longer communication range proves to be indeed beneficial, our method is able to optimise all the parameters effectively even with a severely limited communication radius.
For reference, only 24\% of the robots are within communication range at a 10m radius whereas the percentage increases to 85\% at a 20m radius and yet such a drastic increase only yields insignificant returns in terms of accuracy.

The imposed limit on the communication range also tests the system's asynchronicity. Since an agent cannot communicate at the time of observation if the other agent is too far away, the agent needs to wait for a rendezvous event in order to exchange information.
 Our result hence further highlights that our system is capable of handling asynchronous events.
 
\begin{table}[!t]
\caption{
Evaluation of our method on real-world data. Results include the RMSE ATE of our system
with and without the autocalibration enabled, on UTIAS MR.CLAM dataset 1-4. The noise column indicates whether a noise was artificially added to the sensor calibration to simulate an uncalibrated system.}
\label{table:mrclam}
\vspace{-1em}
\vspace{2mm}
\centering
\begin{tabular}{c|c|cccc|c}
\hline
Noise        &   Auto Calib.  & 1 & 2 & 3 & 4 & Avg. \\ \hline
 \xmark & \xmark & \textbf{0.102} & 0.0976 & \textbf{0.0690} & 0.0712 & 0.0851 \\ 
\xmark    & \cmark    & \textbf{0.102} & \textbf{0.0967} & 0.0706 & \textbf{0.0694} & \textbf{0.0847} \\ \hline
 \cmark    & \xmark & 0.122 & 0.121  & 0.0974 & 0.0855 & 0.107 \\ 
\cmark        & \cmark    & \textbf{0.111} & \textbf{0.120}   & \textbf{0.0825} & \textbf{0.0809} & \textbf{0.0984}  \\ \hline
\end{tabular}
\vspace{-2em}
\vspace{2mm}
\end{table}

\subsubsection{Real-world Experiments}
To verify the applicability of our approach to real-world data, we have performed localisation and auto-calibration using UTIAS MR.CLAM dataset~\cite{Leung:etal:IJRR2011}. As the robots are ground vehicles,  we model them with \SE2 poses and 2D range-bearing observations. We use datasets 1-4, where the landmarks and the robots are randomly scattered. We set $\sigma_B^t = (0.05\text{m}, 0.01\text{m}), \sigma_B^R=5\deg, \sigma_s = (0.08\text{m}, 2\deg)$. The dataset is subsampled at 1s intervals, and we use sliding window-based GBP~\cite{Murai:etal:TRO:2023} with a window size of 30.

As presented in Table~\ref{table:mrclam}, as the robots are calibrated, autocalibration does not yield better RMSE ATE. We simulate an uncalibrated system by artificially adding noise to the sensor calibration with a standard deviation of 0.05m and $10\deg$ for translation and rotation respectively. While this manipulation of calibration is artificial, this data still contains challenging real-world sensor and odometry noise. In such a case, auto-calibration reduces the ATE, indicating that our method has successfully filtered out the biases even with real-sensor data.

\section{Conclusion}
In this paper, we presented a method for online, simultaneous localisation and automatic extrinsic calibration of sensors and observable markers, by building on our previous work Robot Web~\cite{Murai:etal:TRO:2023}.
Our work performs distributed and asynchronous inference on the factor graph using GBP, and we have demonstrated its robustness against large amounts of communication failure, outlying measurements, and restricted communication ranges. 

Distributed and asynchronous properties of GBP offer attractive features for multi-robot systems.
Automatic calibration ensures that the robots require as little maintenance as possible and the accurate localisation provides the basis required for multi-robot interaction.


%

\section*{Acknowledgment}
The authors would like to thank Eric Dexheimer for insightful discussions.




%
\bibliographystyle{IEEEtran}
\bibliography{root}




\end{document}